\documentclass[10pt,journal,compsoc]{IEEEtran}

\usepackage{booktabs}

\ifCLASSOPTIONcompsoc
  \usepackage[nocompress]{cite}
\else
  \usepackage{cite}
\fi

  \usepackage[pdftex]{graphicx}
  \graphicspath{{./figure}}
  \DeclareGraphicsExtensions{.pdf,.jpeg,.png}

\usepackage{bm}
\usepackage{amsmath}
\usepackage{amsfonts,amssymb}
\usepackage{ragged2e}
\usepackage{steinmetz}

\usepackage{amsmath,amsfonts}
\usepackage[ruled]{algorithm2e}
\usepackage{array}
\usepackage[caption=false,font=footnotesize,labelfont=rm,textfont=rm]{subfig}
\usepackage{textcomp}
\usepackage{stfloats}
\usepackage{url}
\usepackage{verbatim}
\usepackage{graphicx}

\usepackage{multirow}
\usepackage{cleveref}
    \crefname{section}{Sec.}{Secs.}
    \Crefname{section}{Section}{Sections}
    \Crefname{table}{Table}{Tables}
    \crefname{table}{Tab.}{Tabs.}
    \Crefname{figure}{Fig.}{Figs.} 
\usepackage[dvipsnames]{xcolor}
\definecolor{red}{rgb}{1,0.6,0.6}
\definecolor{orange}{rgb}{1,0.8,0.6}
\definecolor{yellow}{rgb}{1,1,0.6}

\usepackage{xspace} 
\newcommand{\eg}{e.g.\xspace}
\usepackage{chngcntr}
\usepackage{setspace}

\begin{document}

\title{MixLight: Borrowing the Best of both Spherical Harmonics and Gaussian Models}
%

\author{Xinlong Ji, Fangneng Zhan, Shijian Lu, Shi-Sheng Huang, Hua Huang,~\IEEEmembership{Senior Member,~IEEE}
\IEEEcompsocitemizethanks{
\IEEEcompsocthanksitem Xinlong Ji is with the School of Computing Science \& Technology, Beijing Institute of Technology, Beijing 100081, China. E-mail: jixinlong@bit.edu.cn.
\IEEEcompsocthanksitem Fangneng Zhan is with the Max Planck Institute for Informatics, Saarbrücken 66123, Germany. E-mail: fzhan@mpi-inf.mpg.de.
\IEEEcompsocthanksitem Shijian Lu is with the School of Computer Science and Engineering, Nanyang Technological University, 639798, Singapore. E-mail: shijian.lu@ntu.edu.sg.
\IEEEcompsocthanksitem Shi-Sheng Huang and Hua Huang are with the School of Artificial Intelligence,
Beijing Normal University, Beijing 100875, China. E-mail: \{huangss,
huahuang\}@bnu.edu.cn.

}
\thanks{Hua Huang is the corresponding author.}
}

\markboth{~Vol.~XX, No.~X, 2024}%
{Xinlong Ji, Fangneng Zhan, \MakeLowercase{\textit{et al.}}: MixLight: Borrowing the Best of both Spherical Harmonics and Gaussian Models}


\IEEEtitleabstractindextext{%
\begin{abstract}\justifying
Accurately estimating scene lighting is critical for applications such as mixed reality. Existing works estimate illumination by generating illumination maps or regressing illumination parameters. However, the method of generating illumination maps has poor generalization performance and parametric models such as Spherical Harmonic (SH) and Spherical Gaussian (SG) fall short in capturing high-frequency or low-frequency components. This paper presents MixLight, a joint model that utilizes the complementary characteristics of SH and SG to achieve a more complete illumination representation, which uses SH and SG to capture low-frequency ambient and high-frequency light sources respectively. In addition, a special spherical light source sparsemax (SLSparsemax) module that refers to the position and brightness relationship between spherical light sources is designed to improve their sparsity, which is significant but omitted by prior works. Extensive experiments demonstrate that MixLight surpasses state-of-the-art (SOTA) methods on multiple metrics. 
In addition, experiments on Web Dataset also show that MixLight as a parametric method has better generalization performance than non-parametric methods.
\end{abstract}

}

\maketitle

\IEEEdisplaynontitleabstractindextext

\IEEEpeerreviewmaketitle

\section{Introduction}\label{sec:intro}

\noindent
\IEEEPARstart{R}{ecovering} high-dynamic-range (HDR) illumination from a single limited field of view (FOV) image is a challenging yet attractive task. 
This task holds significant relevance across various domains within computer graphics, encompassing mixed reality, HDR relighting, and virtual characters insertion.
Addressing this challenging problem entails confronting three primary challenges.
First, inferring panoramic illumination from a image of limited FOV poses a prominent difficulty. 
Second, recovering HDR illumination from low dynamic range (LDR) observation falls into an ill-posed problems.
Third, the observed images are the result of intricate interplay between geometry, material, and illumination, leading to substantial ambiguity for lighting estimation.


\begin{figure}[t]
  \centering
  \subfloat[Ambient Light estimation]{\includegraphics[width=0.48\linewidth]{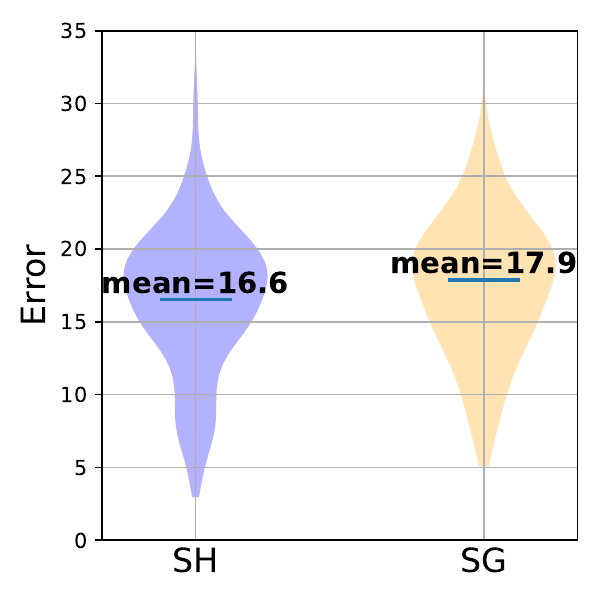}\label{fig:short-ambient}}
  \hfil
  \subfloat[Light Sources estimation]{\includegraphics[width=0.48\linewidth]{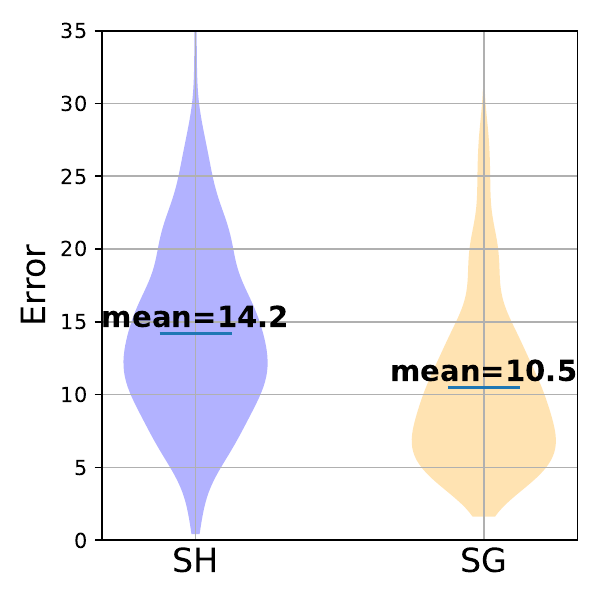}\label{fig:short-light}}
  \caption{Two violin figures illustrate the distinct advantages of SH and SG in illumination estimation tasks. The objective is to examine the potential disparities between SH and SG functions in representing various light components. In the first scenario, SH and SG with the same parameter sizes are used to represent \textit{ambient light}. Two networks are trained to predict the parameters of SH and SG respectively. The predicted ambient light is then employed to render spheres (depicted in~\Cref{fig:speres}) on the test set. Prediction accuracy is evaluated by calculating the error between the predicted rendering result and the real rendering image. The resulting errors from all test set samples are visualized in the violin figure~\Cref{fig:short-ambient}, clearly displaying the distribution and average of the errors. In the second scenario, SH and SG with similar parameters are used to represent the \textit{light source}. The network is trained, and the rendering error is then plotted in~\Cref{fig:short-light}. For further details about the experiment's design, refer to the supplementary file.
  }
  \label{fig:sh_vs_sg}
\end{figure}

Prior works~\cite{debevec1998rendering,li2013capturing,knorr2014real,georgoulis2017around,barron2013intrinsic,gruber2012real,bin2017static} mitigate this ill-posedness by incorporating additional information; however, these approaches often lack user-friendliness, such as requiring obtaining the depth map~\cite{barron2013intrinsic,gruber2012real} or multi-view pictures~\cite{li2013capturing,bin2017static}. Recent works have leaned towards leveraging neural networks to predict illumination directly from a single image, either by generating illumination maps~\cite{gardner2017learning,wang2022stylelight,zhan2021emlight,zhan2022gmlight,dastjerdi2023everlight,weber2022editable} or by regressing illumination parameters~\cite{gardner2019deep,zhan2021emlight,zhan2022gmlight,li2019deep,zhan2021sparse,garon2019fast,xu2020real,dastjerdi2023everlight,weber2022editable}. 
Illumination maps, which are panoramic images containing illumination information in all directions, offer the potential to fully represent illumination. However, their high-dimensional nature often results in stability issues and poor generalization performance for lighting estimation.
On the other hand, the regression-based methods parameterize the illumination with basis functions such as SH~\cite{garon2019fast,cheng2018learning,xu2020real} and SG~\cite{gardner2019deep,zhan2021emlight,zhan2022gmlight,li2019deep,dastjerdi2023everlight,weber2022editable}, followed by a network to regress the illumination parameters of lower dimensions. 
These methods have better generalization performance but struggle to capture either high-frequency or low-frequency characteristics, leading to incomplete representation.

\begin{figure*}[t]
\includegraphics[width=1.0\textwidth]{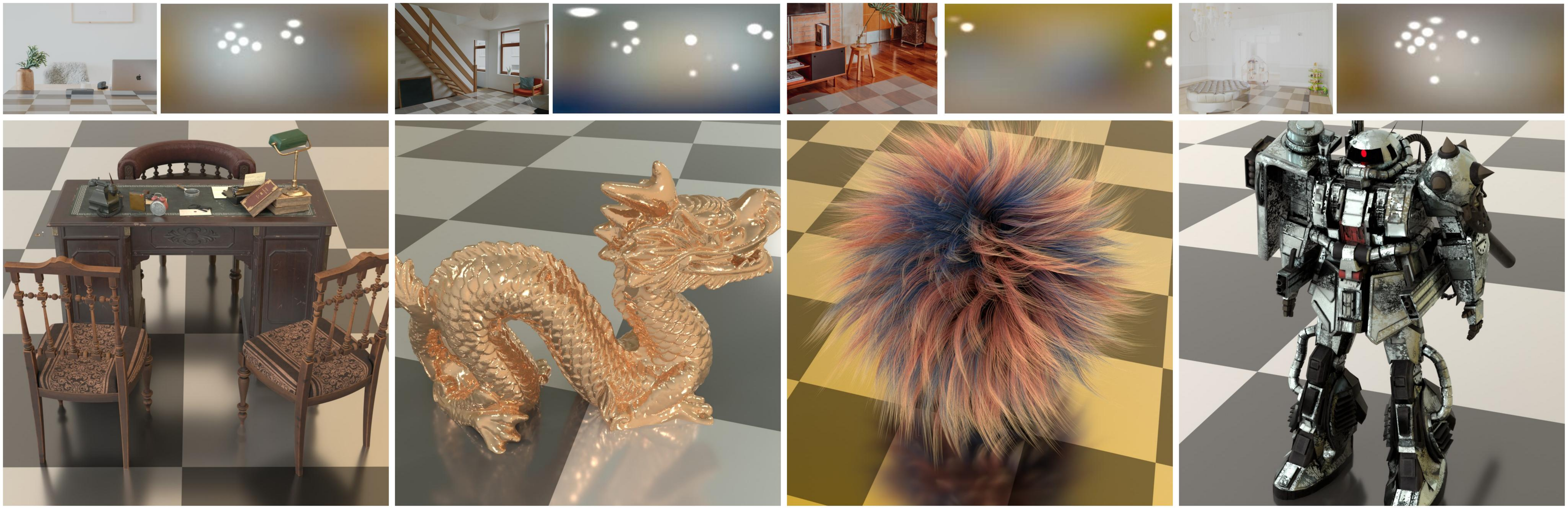}
\vskip 5pt 
\caption{ The proposed MixLight estimates illumination and re-illuminates multiple virtual objects (in the second row). MixLight estimates low-dimensional lighting parameters that can be visualized as illumination maps (at the top right of each example) from limited field-of-view pictures (at the top left of each example).}
\end{figure*}

Specifically, \Cref{fig:sh_vs_sg} shows that SH is better suited for capturing low-frequency ambient light (with a less error in \Cref{fig:short-ambient}), while SG is more suitable for capturing high-frequency light sources (with a less error in \Cref{fig:short-light}). Therefore, the joint representation of illumination by utilizing the complementary advantages of SH and SG in the frequency-domain can lead to a more accurate representation of illumination information.

Inspired by this, a SH, SG joint model of illumination representation called MixLight is proposed. An HDR illumination map can be divided into two parts: ambient light and light source using a simple brightness threshold segmentation method~\cite{zhan2021emlight}. Among them, the ambient light is mainly the low-frequency illumination information. The light source is a steep peak in brightness, which corresponds to the high-frequency part of the illumination information. MixLight uses a neural network to regress both SH and SG parameters simultaneously, where SH represents the low-frequency ambient light part and SG represents the high-frequency light source part.

In practice, real-world light sources often exhibit a sparse characteristic, typically confined to a relatively small yet variable number. Nevertheless, prior SG methods either overlook this sparsity~\cite{zhan2021emlight,zhan2022gmlight} by generating extraneous tiny light sources, or trap in oversimplified settings by featuring a fixed number of light sources~\cite{gardner2019deep,dastjerdi2023everlight,weber2022editable}.
Inspired by the sparsemax theory~\cite{martins2016softmax,laha2018controllable}, this paper designs the \textit{Spherical Light Source Sparsemax} (SLSparsemax) mechanism to impose sparsity constraints on light sources at the neural network level. SLSparsemax considers the positions and brightness relationships among light sources on the sphere, calculates the credibility of light source predictions, and adaptively determines the number of sparse light sources.

The contributions of this work are summarized as follows:

\begin{enumerate}
    \item A SH, SG joint model named MixLight is proposed to represent illumination more completely and accurately with a few parameters.
    \item The spherical light source sparsemax (SLSparsemax) mechanism is designed to impose sparsity constraints on the estimation of light sources.
    \item Experiments on several datasets show that MixLight outperforms several SOTA methods in prediction accuracy and has good generalization ability.
\end{enumerate}

\begin{figure*}[t]
    \centering
    \includegraphics[width=1.0\linewidth]{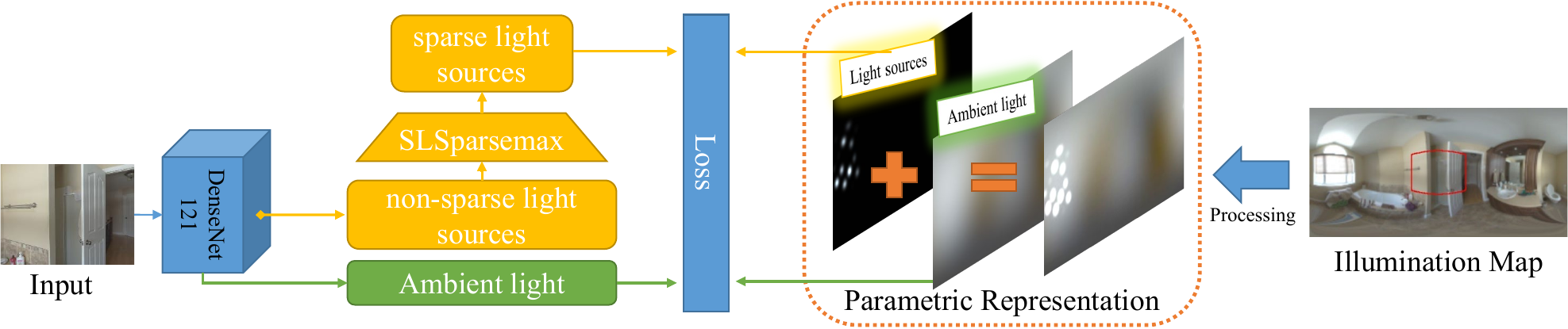} 
    \caption{MixLight parameters decomposition and estimation. In the right half of the figure, an illumination map is separated as the light sources and the ambient light component, then decomposed to true values of MixLight parameters (including SG and SH parameters). 
    The left half describes the parameter regression process. Specifically, the MixLight model uses an SLSparsemax activation function layer in the network to enforce the sparsity of light source parameters. Consistent with \cite{zhan2021emlight,zhan2022gmlight,garon2019fast,gardner2019deep}, MixLight utilizes DenseNet121 as the backbone network. The input of the network is cropped from the illumination map (corresponding to the area in the red box).}
    \label{fig:MixLightOverview}
\end{figure*}
\section{Related Works}

\noindent
Illumination estimation is a classic and challenging task in the fields of computer vision and computer graphics. The panoramic HDR illumination estimated from a single image can be directly used as illumination conditions for many subsequent applications such as  the virtual
characters insertion or relighting~\cite{mei2023lightpainter,nestmeyer2020learning,pandey2021total,sun2019single,wang2020single,yeh2022learning,zhang2021neural-CVPR,zhang2021neural-TOG,meka2019deep,kim2024switchlight}.
Early research efforts attempt to introduce more information to alleviate the ill-posedness, such as using landmarks in the scene to obtain partial geometric (or material) information~\cite{debevec1998rendering,li2013capturing,knorr2014real,georgoulis2017around,weber2018learning,yu2023accidental}, or directly obtaining scene geometry from depth maps~\cite{barron2013intrinsic,gruber2012real} and multi-view pictures~\cite{li2013capturing,bin2017static}. However, this introduction of additional information is not user-friendly.

In recent years, most illumination estimation works are based on neural networks, which generate illumination maps~\cite{gardner2017learning,zhan2021emlight,wang2022stylelight,zhan2022gmlight,dastjerdi2023everlight} or regress 
parameters~\cite{gardner2019deep,zhan2021emlight,zhan2022gmlight,li2019deep,zhan2021sparse,garon2019fast,xu2020real,cheng2018learning,dastjerdi2023everlight,weber2022editable}, no longer relying on additional information. To generate high-dimensional illumination maps, a large amount of real HDR training data is required to ensure the generalization performance and the realism of the prediction. However, constructing such large-scale realistic datasets that currently lacks consumes considerable time and effort. Therefore, Gardner et al.~\cite{gardner2017learning} train a CNN mainly on a large-scale synthetic dataset and then fine-tuned light source intensity on a small-scale real dataset. More recent works \cite{zhan2021emlight,wang2022stylelight,weber2022editable,dastjerdi2023everlight} rely on the powerful generative adversarial network (GAN) to generate illumination maps with realistic details. Zhan et al.~\cite{zhan2021emlight} use the regressed parameters of SG to guide GAN~\cite{park2019semantic} generating, similar to another two hybrid models~\cite{weber2022editable,dastjerdi2023everlight}, while Wang et al.~\cite{wang2022stylelight} directly generate illumination maps with scene style encoded in styleGAN~\cite{karras2020analyzing}. However, GANs are trained only on the limited-scale real HDR dataset, resulting in overfitting to the training set and poor generalization performance.

Parameters regressing methods use parameterized representations such as SH~\cite{cheng2018learning,xu2020real,garon2019fast} and SG~\cite{gardner2019deep,zhan2021emlight,zhan2022gmlight,li2019deep}. As SH supports low overhead, fast and differentiable illumination maps reconstruction and object rendering, SH based methods are introduced in resource-limited mobile devices~\cite{cheng2018learning,xu2020real,garon2019fast}, as well as reconstruction loss~\cite{xu2020real} and rendering loss~\cite{cheng2018learning} that improves the quality of predictions. However, limited to representation defects in the frequency domain (shown in~\Cref{fig:sh_vs_sg}), SH methods represent high-frequency illumination inaccurately. Oppositely, SG methods are poor in low-frequency representation, but also suffer from inaccuracy in high-frequency representation as their thoughtless about the sparsity of light sources. Gardner~et al.~\cite{gardner2019deep} oversimplify the number of light sources to a fixed number (\eg, 3), which cannot accurately represent the varying number and shape of light sources in the real world. Zhan et al.~\cite{zhan2021emlight} achieve variable SG light sources using spherical distribution but fail to remove numerous small distribution values in the predicted results, which is also not sparse, not consistent with reality, and misleads the subsequent generation. 

As a method that generates high-dimensional illumination maps requires a large-scale realistic dataset currently unavailable to ensure generalization performance, using low-dimensional parameterized representations is a good alternative. Note that parametric representations can better connect with many existing applications~\cite{zhu2023facescape,feng2021learning,chen2020neural,zhou2019deep} that choose parametric illumination for a more efficient rendering. However, previous works typically use either SH or SG independently, which fail to accurately represent both high-frequency and low-frequency illumination simultaneously. This paper introduces MixLight, which combines the strengths of SH and SG, to respectively represent low-frequency ambient light and high-frequency light sources, achieving a lower-dimensional yet more complete representation. Against the non-sparsity that weakens the accuracy of high-frequency light source prediction, SLSparsemax is designed, which leverages the positional and intensity information of spherical light sources to help the network output adaptively sparse light sources.


\section{Proposed Method}

\noindent
The proposed MixLight combines the complementary characteristics of SH and SG, using them to represent the ambient light and the light sources respectively, as illustrated in \Cref{fig:MixLightOverview}. SLSparsemax is designed in regressing network to improve the accuracy of high-frequency illumination prediction. The following sections introduce the methods used for the MixLight parameters decomposition (\Cref{subsec:decomposition}), the design of SLSparsemax (\Cref{subsec:slsparsemax}), and details about loss functions (\Cref{subsec:lossFunction}).

\subsection{Parameters Decomposition}\label{subsec:decomposition}
\noindent
As visualized in the right half of \Cref{fig:MixLightOverview}, the illumination map $I$ is separated into the ambient light $I_A$ and the light source regions $I_L$ firstly, which are further parameterized by Spherical Harmonics and Spherical Gaussian, respectively.

\textbf{Separation.} 
In the real world, the energy of light sources is mainly located in the high-frequency domain while ambient light is comparatively low-frequency. This distinction arises from the fact that 
the light source exhibits sharp changes in brightness, whereas ambient light changes more gradually. Separating light sources from ambient light effectively separates high and low-frequency light, facilitating subsequent parameters decomposition processes.
In addition to the frequency differences, there is a huge brightness difference in that, light sources are always much brighter than other objects that comprise ambient light in the scene. HDR format illumination maps can capture such a huge brightness difference, where the above two light types can be effectively separated by using a 5\% brightness threshold, as proposed in~\cite{zhan2021emlight}. This method involves identifying the top 5\% brightest pixels as light sources and categorizing the others as ambient light.


\textbf{SH Parameters Decomposition. }In the field of lighting estimation, SH is an important method for representing illumination \cite{cheng2018learning,garon2019fast,xu2020real}. Similar to the Fourier transform, SH can use a set of orthogonal spherical basis and their corresponding coefficients to approximate illumination~\cite{frolova2004accuracy,ramamoorthi2002analytic,ramamoorthi2006modeling}, especially to represent the ambient light $I_A$ in this paper, as shown in \Cref{eq:lowSH2ambient}:


\begin{equation}
I_A(c,\rho) \approx \sum_{k=0}^{K}\sum_{m=-k}^k  A_{m,c}^k B_m^k(\rho)
  \label{eq:lowSH2ambient}
\end{equation}

Here, $c$ represents the channel of RGB format $I_A$, where $c$ can be 1, 2, and 3, corresponding to red, green, and blue channels, respectively. $\rho = (\phi, \theta)$ represents the spherical coordinate angles defined on the spherical image domain $\Omega$, where $\phi$ is the azimuth angle and $\theta$ is the elevation angle. $B_m^k$ denotes the spherical harmonic basis function with order $k$ (ranging from 0 to $K$) and degree $m$ (ranging from -$k$ to $k$), and $A_{m,c}^k$ represents the corresponding coefficients for the channel $c$. $K$ is the upper bound of $k$. 

As the order $k$ increases, SH can capture information from low frequencies to high frequencies, resulting in a more accurate representation~\cite{frolova2004accuracy,ramamoorthi2002analytic,ramamoorthi2006modeling}. However, SH should only be used to represent low-frequency information considering its representation defects, and the scale of the $A_{m,c}^k$ also rapidly expands with increasing order. Thus, blindly raising the upper boundary of $k$ just leads to SH losing its advantage of parameter sparsity, without much real gain. It is common to only use low-order SH to approximate illumination, setting $K$ as a small value.

As the inverse process of \Cref{eq:lowSH2ambient}, the SH coefficients $A_{m,c}^k$ can be obtained by projecting the ambient light $I_A$ onto the SH basis: 
\begin{equation}
    A_{m,c}^k = \frac{4\pi}{wh} \sum_{\rho}^{\Omega}  I_A(c,\rho) B_m^k(\rho) 
    \label{eq:ambient2SH}
\end{equation}
Here, $w$, $h$, and their product $wh$ represent the width, the height, and the area of the ambient light $I_A$.

\textbf{SG Parameters Decomposition.} Similar to the approach in \cite{zhan2021emlight}, the light source part $I_L$ is represented using SG on $N$ evenly distributed anchor points. The intensity of the light source is denoted as $E$, its color is represented by $R$, and the energy distribution of light sources is modeled using the light distribution $P$. The method of seeking the SG parameters is also basically similar to \cite{zhan2021emlight}. 
To explain this process more clearly, details are provided here.

To calculate the light source intensity $E$, sum up the pixel values in each channel, yielding:
\begin{equation}
 T_{c} = \sum_{\rho}^{\Omega} I_{L}(c,\rho), \quad c=1,2,3 
\end{equation}
$T_{c}$ represents the sum of the pixel values in the $c$ channel.

The total intensity of the light source is the L2 norm of the vector formed by $T_1$, $T_2$ and $T_3$:
\begin{equation}
E = \left|\left| \left( T_1, T_2, T_3 \right) \right|\right|_2^2
\end{equation}

The overall color of the light source is represented by the color ratios $R$:
\begin{equation}
R = \frac{(T_{1}, T_{2}, T_{3})}{E}
\end{equation}

Next, the pixel values within the light source region are assigned to the nearest anchor point, and the energy distribution on $N$ anchor points is computed as $P$. 

As a reverse process of parameter decomposition, light sources can be reconstituted in the form of spherical Gaussian maps, as employed in Zhan et al.~\cite{zhan2021emlight}. However, it is important to note that the \textit{Gaussian Map} reconstructed by Zhan is not entirely accurate. The reconstructed Gaussian map mainly approximates the distribution of light, serving primarily as a rough guideline for subsequent generators, without providing an accurate intensity measure.

Different from Zhan et al.~\cite{zhan2021emlight} but similar to the idea of~\cite{li2019deep,weber2022editable,dastjerdi2023everlight}, the Gaussian map generated by MixLight directly serves as \textit{final result} of the light source part estimation, which must be numerically accurate in intensity. Therefore, a normalization term $q$ is multiplied in the Gaussian map reconstruction function to ensure that the integral of a single Gaussian kernel is kept as 1 (before multiplied by the amplitude value), which aims to keep the pixel value sum of pixels that assigned to an anchor point in $I_L$ equal to the integral of the corresponding SG pixel values in  Gaussian map. 
Finally, the Gaussian map reconstruction formula is written as: 
\begin{equation}
I_{L} \approx \sum_{i=1}^{N} v_{i}e^{\frac{d_{i}u-1}{s}}q
\end{equation}
$N$ is the number of anchor points, $v_i$ denotes the RGB values of the $i$-th anchor points, which is the product of the light distribution, intensity, and color ratios on that anchor point (namely $v_i = P_i E R$), $d_i$ is the direction of an anchor point (predefined by the Vogel method \cite{vogel1979better}), $u$ is the unit vector giving a direction on the sphere, $s$ is the angular size of the Gaussian kernel. 

Normalization term $q$ is the reciprocal of the integral value of the spherical Gaussian and is dependent on the angular size $s$ and the sphere radius $r$:
\begin{equation}
q = \frac{1}{{2 \pi s r^{2} \left(1 - e^{-2s}\right)}}
\end{equation}


\textbf{Discussion.} Why use SH to represent low-frequency ambient light instead of using other functions on the sphere, such as SG? After all, by adjusting the angular size of the SG, it can be made smoother as well to represent low-frequency information. In the comparison experiment (\Cref{fig:sh_vs_sg}) illustrated in the \textit{Introduction} section, the angular size of SG is indeed carefully adjusted to produce two versions of SG. One sharp version is physically closer to the light source, which is compared with SH in the light source estimation task; another smooth version of SG that is physically closer to ambient light is used for comparison with SH in ambient light estimation tasks. More details about this adjustment can be found in the supplementary file. In this experiment, the parameters of SG were carefully set for the sake of fairness of comparison. But this also reflects a problem. When SG is used to represent lighting, many aspects of it (such as the value of the angular size, the number of SG kernels, whether it is a position-fixed model to predict the amplitude like \cite{zhan2021emlight,zhan2022gmlight} or a number-fixed model to predict the position and amplitude simultaneously like \cite{dastjerdi2023everlight,gardner2019deep}) all requires someone with specialized knowledge to determine, and the process involves too many variable parameters and is often very cumbersome and uncertain. In contrast, SH only needs to determine which order to use, which is simple but efficient. In addition, previous works ~\cite{frolova2004accuracy,ramamoorthi2002analytic,ramamoorthi2006modeling} have fully studied the usage of SH to represent low-frequency illumination information, where the 2-order SH has been proven to be very effective. Therefore, even users who lack professional knowledge can obtain a better lighting representation effect (refer to \Cref{fig:short-ambient}) by simply setting the order to 2. In general, except for excellent accuracy in representing low-frequency light, SH has far fewer parameters to adjust and is easy to set, which is more user-friendly and efficient.

\subsection{SLSparsemax}\label{subsec:slsparsemax}

\noindent
Light sources in the real world are sparse. In this paper, the sparsity of light sources refers to: in a scene, there are usually only a few light sources, and the number of light sources is variable. Previous works on SG do not effectively constrain the sparsity of the predicted results, which leads to inaccurate predictions. 


Sparsemax, as described in \cite{martins2016softmax}, is an activation function utilized as the final layer in the neural network to improve the sparsity of the output. It has been applied in multi-class classification problems \cite{laha2018controllable}. When used in illumination estimation, sparsemax tends to retain several of the brightest light sources while filtering out others, which helps create a sparsity relative to individual SG light sources. Further details on sparsemax can be found in the supplementary file and~\cite{martins2016softmax}.

As a general method, the original sparsemax fails to utilize additional information in illumination estimation issues. 
In the real world, light sources usually have different shapes (\eg, patches, bars, points) and varying areas, implying useful prior information for lighting estimation.
However, the original SG light source is point-shaped. Therefore, when simulating light sources using SG, it is encouraged to connect local point SG light sources on the spherical surface to simulate the patch-like and bar-like light sources in the real world. 
To this end,
sparsity should be maintained between several SG clusters, rather than just between SG individuals. In summary, SLSparsemax encourages ``local clustering, global sparsity'' for SG light sources.



How can the SG point light on the sphere form a ``locally clustered, globally sparse'' distribution? The classical intrinsic decomposition task~\cite{zhao2012closed,grosse2009ground,sengupta2018sfsnet,meka2021real,wang2019underexposed} seems to have similar requirements. The intrinsic decomposition task decomposes a picture into a reflectance layer and a shading layer, where the reflectance layer represents the color of the object itself, and the shading layer represents the effect of light on the object. The reflectance layer is usually assumed to consist of a limited number of color blocks, and the pixels within a block have the same color, but the blocks differ from one another. This is also a prior of local clustering and global sparsity. To satisfy this prior, the chroma or brightness similarity of a pixel with adjacent pixels is often measured, which is then incorporated into iterative optimization to help acquire the reflectance value. Inspired by this, the brightness similarity of one spherical light source with its neighborhood is used in SLSparsemax, where light sources more similar to neighbors have a better chance of being preserved. In this setting, if two light sources have the same brightness in the initial prediction result, the one with a more similar brightness to the neighborhood will be preferentially retained to encourage the network to eventually output a ``locally clustered, globally sparse'' light source prediction. Even if the wrong light source is retained, it will also be punished by the supervision of subsequent loss function, thus improving the prediction accuracy of the network.

%
\begin{algorithm}[h] 
\doublespacing 
\SetAlgoLined 
\caption{SLSparsemax} 
\label{alg:SLSparsemax}
\SetKwInOut{Input}{Input} 
\SetKwInOut{Output}{Output} 
\Input{$P$} 

Normalize $P^{norm}= P - \max \left\{P\right\}$ \\
Calculate $P^{cred}$ \text{s.t.} \\
\Indp
\text{$P_i^{simi} = \exp\left(-|P^{norm}_i - \frac{1}{|\Gamma(i)|} \sum \limits_{j \in \Gamma(i)}P^{norm}_j| \right)$}\\
\text{$P_i^{cred} = \frac{P_i^{norm}}{P_i^{simi}}$}\\
\Indm
Sort $P$ as $P^{cred}(1) \geq \dots \geq P^{cred}(N)$ \\

Find $\kappa(P):=$ \\
\Indp
\text{$\max \left\{ \kappa \in [N] \mid 1 + \kappa \min \limits_{1 \leq i \leq \kappa}\{P(i)\}> \sum \limits_{j \leq \kappa} P(j) \right\}$}\\
\Indm
Define $\tau (P) = \frac{\left(\sum_{j \leq \kappa(P)} P(j) \right)-1 }{\kappa(P)}$ \\

\Output{$P^{sparse}$ \text{s.t.} $P^{sparse}_i = \left[ P_i - \tau (P) \right]_{+}$}

\end{algorithm}


As depicted in \Cref{alg:SLSparsemax}, SLSparsemax takes the initially predicted non-sparse light distribution $P$ as input. Each element in $P$ represents the brightness probability of a corresponding light source located on the spherical surface. To effectively filter out small light sources and locally cluster light sources, the \textit{credibility} $P_i^{cred}$ of each SG light source is computed based on its normalized brightness probability $P_i^{norm}$ and its \textit{similarity} $P_i^{simi}$ (in terms of brightness) with neighboring light sources in its neighborhood $\Gamma(i)$. Subsequently, the input $P$ is sorted in descending order of the corresponding credibility $P_i^{simi}$. Through a convex optimization process, a threshold $\tau (P)$ is determined to retain the top $\kappa(P)$ most reliable light sources, thereby producing the sparse output $P^{sparse}$. In contrast to the original sparsemax~\cite{martins2016softmax} which focuses solely on the brightness of light sources, SLSparsemax additionally takes into account the brightness similarity between the light source and its neighborhood, where higher similarity results in enhanced gains in credibility, and ultimately a higher chance of being retained.

\subsection{Loss Functions}\label{subsec:lossFunction}


\noindent
For the SG section, MixLight adopts some effective SG-related loss functions that have been proven useful in~\cite{zhan2021emlight}. Same with~\cite{zhan2021emlight}, the original L2 loss is used to constrain the distribution of lighting sources $P$, the intensity $E$ and the color ratio $R$, as well as their SML loss for $P$. These loss functions can be found directly in Zhan et al.~\cite{zhan2021emlight}.
In addition to these, a \textit{masked L1 loss} is devised for the SG parameter $P$, in conjunction with SLSparsemax, to ensure accurate estimation of the light sources on the lower hemisphere.


The original sparsemax is a general method and not directly applicable to the lighting estimation problem. Unlike multi-class classification problems where sparsemax has been introduced and improved sparsity, the light distribution $P$ in lighting estimation problems is biased, with the majority of light sources distributed in the upper hemisphere of the spherical space. In this case, directly introducing the original sparsemax in the neural network will cause the network to inadequately learn the light distribution in the lower hemisphere of the spherical space, resulting in the network predicting numerous small light sources in the lower hemisphere where there are originally very few light sources. SLSparsemax also failed when facing this special light distribution-biased problem. 
To solve this problem, a masked L1 loss function specifically designed for light distributions $P$ is introduced to penalize the predicted tiny light sources in some places where there is no light source in the GT, which mainly occurs in the lower hemisphere. In other words, the masked L1 loss function $L_{masked-L1}$ will penalize tiny light sources that occur (have non-zero value) in predicted $\widehat{P}$ but do not occur in ground truth $P$. This loss function compensates for the inadequacy of the original sparsemax method in handling the bias in data.

\begin{equation}
\label{eq:maskedL1}
L_{masked-L1} = \lVert M^P \odot (\widehat{P} - P) \rVert_1
\end{equation}
$M^P$ is a mask with the same shape as $P$. Where $P$ is 0, the corresponding position in $M$ is 1; where $P$ is non-zero, the corresponding position in $M$ is 0. $\odot$ means Mask operation.

Regarding the SH parameters, MixLight mainly considers constraints on the SH coefficients themselves, constraints on the SH reconstruction results (used in \cite{xu2020real}), and constraints on the SH rendering results (used in \cite{cheng2018learning}). 

The loss function for SH coefficients is defined as: 

\begin{equation}
L_{\text{SH-co}} =  \sum_{c=1}^{3} \sum_{k=0}^{K} \left( \frac{1}{2k+1} \sum_{m=-k}^{k} \left( \widehat{A}_{m,c}^k - A_{m,c}^k \right)^2 \right) 
\end{equation}

However, achieving a low error in the spherical harmonic (SH) coefficients does not necessarily ensure accurate light prediction. Even minor variations in the SH coefficients can result in significant alterations to both the reconstructed lighting map and the rendered outcomes. Therefore, imposing direct constraints on both the reconstructed lighting map and the rendering results can prove beneficial~\cite{xu2020real,cheng2018learning}.

Thanks to the differentiable nature of SH reconstruction \cite{xu2020real} for light map and rendering \cite{cheng2018learning}, the SH reconstruction loss function and the rendering loss function can be directly constructed and optimized by backpropagation to update the network parameters. The SH reconstruction loss has been confirmed to contribute to the stability of the training process \cite{xu2020real}, which is defined in \Cref{eq:SHReconstructLoss}:

\begin{equation}
\small
\label{eq:SHReconstructLoss}
 \frac{1}{3wh} \sum_{c,\rho} \cos\theta \left(  \sum_{m,k} \left( \widehat{A}_{m,c}^k B_{m}^k\left(\rho\right) - A_{m,c}^k B_{m}^k\left(\rho\right) \right) \right)^{2} 
\end{equation}

Similar to \cite{gkitsas2020deep}, importance weighting $\cos{\theta}$ is also applied to the reconstructed lighting map in \Cref{eq:SHReconstructLoss}. In spherical space, higher dimensions occupy smaller proportions of the sphere. Therefore, when calculating pixel errors on the lighting map, weights related to the elevation angle should be used to weaken the influence near the poles and increase the influence near the equator.

As to SH rendering Loss, Cheng et al.~\cite{cheng2018learning} randomly selected a few objects from several prepared objects for rendering during training. However, rendering losses defined on specific objects are not widely representative. In fact, any illumination values in any rendered objects come from the ``irradiance environment map''\cite{ramamoorthi2001efficient}, which preserves the irradiance for all orientations, while the rendering result of a specific object only includes a subset of the irradiance. The former is a superset (or the complete set) of the latter. Therefore, constraining the loss on the irradiance environment map indirectly constrains the loss of all potential objects to be rendered. 

Given predicted SH coefficients $\widehat{A}$ and ground truth $A$, the SH rendering loss $L_{\text{SH-rd}}$ is given by \Cref{eq:SHRenderLoss}, with a function \textbf{shRender}$(\cdot)$ helping to calculate irradiance environment map from coefficients (learn more in \cite{ramamoorthi2001efficient,cheng2018learning}).
\begin{equation}
\small
\label{eq:SHRenderLoss}
\frac{1}{3wh}  \sum_{c,\rho} \cos\theta \left(\text{shRender}(\widehat{A}) - \text{shRender}(A)\right)^{2}
\end{equation}

\section{Experiments}

\subsection{Dataset}

\noindent
To train and evaluate MixLight, the publicly available Laval Indoor HDR Dataset~\cite{gardner2017learning} is utilized. This dataset comprises 2100 HDR illumination maps. Following the data preprocessing approach of~\cite{gardner2017learning,zhan2021emlight}, each HDR panorama is cropped and warped eight times, resulting in a total of 19,556 training pairs. Similar to the method in~\cite{zhan2021emlight}, 200 training pairs are randomly selected as the test set, while the remaining 19,356 pairs are used as the training set.

Most prior works only assess their methods on the Laval Indoor HDR Dataset. However, this paper introduces an additional \textbf{Web Dataset} for testing. The images in the Web Dataset are collected from the internet and have \textit{no intersection} with the Laval Indoor HDR Dataset, ensuring that previous SOTA methods have not reached them. This approach aims to provide a fairer evaluation. Meanwhile, comparing a method's prediction errors on the Web Dataset and the Laval Indoor HDR Dataset allows for the assessment of its generalization performance. Following the same cropping and warping process, the Web Dataset comprises a total of 200 samples for testing.
There are two reasons for performing additional tests on the Web Dataset:
\begin{itemize}
    \item For one thing, generative methods based on GANs have been criticized for their poor generalization performance \cite{zhan2021sparse}. When the training set and test set are split from one limited in-scale dataset (\eg, Laval Indoor HDR Dataset) with a specific scene style, these generative methods are able to memorize such delicate scene style due to their large-scale network parameter, and thus generate delicate illumination maps when tested, performing better than others. However, they may suffer from the overfitting to such a specific dataset style, and perform worse once the test samples are randomly collected from wild scenes with significant style differences. To validate this claim, a dataset having a random style is required.
    \item For another, as some SOTA methods are not open-sourced and all compared methods are trained on the Laval Indoor HDR Dataset, it is difficult to ensure their training sets do not overlap with the test set used in this paper. While, constructing an additional Web Dataset ensures that it does not overlap with the training sets used by all methods, thus providing a fairer evaluation result.
\end{itemize}

\subsection{Training Settings}
\noindent
Following \cite{gardner2019deep,gardner2017learning,zhan2021emlight}, the size of the illumination map is 128x256, and the size of the limited FOV image that serves as the input is 192x256. 
In line with the best practice outlined in \cite{zhan2021emlight}, a 5\% threshold is used to separate ambient light from light sources. Moreover, the number of anchor points $N$ is set to 128, and the angular size $s$ is set to 0.0025, which constructs a sharp SG to represent the sharp light sources. For the setting of $N$, note that in the experiment conducted by Zhan et al.\cite{zhan2021emlight}, the prediction performance drops slightly when 64 instead of 128 anchor points are used, and increasing anchor points to 196 doesn’t improve the performance obviously. They conjecture that the larger number of parameters with 196 anchor points affects the regression accuracy negatively, choosing $N=128$ as the best setting.

Additionally, research by \cite{ramamoorthi2001efficient} shows that using 2nd order SH is sufficient for representing low-frequency ambient light (with an error below 1\% when rendering Lambertian objects). Therefore, we use 2nd order SH to represent the ambient light, setting $K$ to 2. We found that reducing the SH parameter number will lead to the degeneration of representation capabilities, while increasing their number will make the regression task more challenging, leading to unstable training.

As to SLSparsemax, 6 closest points are considered to be in the neighborhood of an SG light source. We have also tried multiplying or dividing the components in the brackets of \textit{credibility} by 10 and finally found that the current configuration is the best. The proposed MixLight is implemented using the PyTorch framework. Adam optimizer with a learning rate decay mechanism (initial learning rate of 0.0001) is used during training. It has been trained 130 epochs with a batch size of 32, on one NVIDIA GeForce RTX 3090 GPU with 24GB of memory.

\begin{figure}[!ht]
  \centering
  \subfloat[Gray Diffuse]{
    \includegraphics[width=0.3\linewidth]{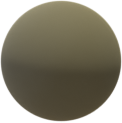}
    \label{fig:short-a}}
  \hfil
    \subfloat[Matte Silver]{
    \includegraphics[width=0.3\linewidth]{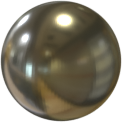}
    \label{fig:short-b}}
    \hfil
        \subfloat[Mirror Silver]{
    \includegraphics[width=0.3\linewidth]{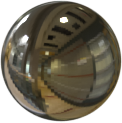}
    \label{fig:short-c}}
  \caption{The scenes used in evaluations consist of three spheres with different materials including diffuse gray, matte silver and mirror silver.}
  \label{fig:speres}
\end{figure}

\subsection{Evaluation Method and Metric}

\noindent
Like several previous works \cite{zhan2021emlight,wang2022stylelight,zhan2022gmlight,gardner2019deep}, MixLight undergoes both quantitative and qualitative evaluation.

For quantitative evaluation, all the compared methods estimate the illumination of samples from both the test set of the Laval Indoor HDR Dataset and the Web Dataset. The predicted illumination is then used to render three spheres, as shown in~\Cref{fig:speres}, using the Blender rendering engine~\cite{hess2013blender}. The rendered results are then assessed against the GT using commonly used evaluation metrics, including RMSE and si-RMSE~\cite{grosse2009ground}, which focus on the estimated light intensity and light directions (or shadings), respectively.

In qualitative evaluation, the predicted results are first visualized as illumination maps. 
Furtherly, similar 3D scenes, as used by \cite{zhan2021emlight,wang2022stylelight}, each containing a virtual object and a background image, are rendered. The accuracy of illumination estimation can then be reflected based on the visual consistency of the rendered object with the scene. 

Note that the actual predicted result is in HDR form, but only the LDR form picture can be displayed in the paper. However, LDR results do not show the true luminance estimates for each method because the tone mapping process adjusts the dynamic range of all images to the best viewing luminance. Furthermore, showing only the LDR light map would also make the observer psychologically biased toward the generative approach, which is unfair because the generative approach uses a higher-dimensional representation. For this reason, three spheres rendered during quantitative evaluation are shown along with the LDR form illumination maps, which reflect the true luminance and reduce potential unfairness.

\begin{table*}[t]
  \centering
    \caption{Comparison of MixLight with several SOTA lighting estimation methods on two datasets. The evaluation metrics include the widely used RMSE, si-RMSE. D, S, M denote a diffuse, a matte silver and a mirror material of the rendered objects, respectively.}
  \label{tab:Quantitative}
\begin{tabular}{c|ccc|ccc|ccc|ccc}
\hline
 & \multicolumn{3}{c|}{StyleLight} & \multicolumn{3}{c|}{EMLight} & \multicolumn{3}{c|}{Gardner19}& \multicolumn{3}{c}{MixLight}\\ \cline{2-13} 
  Metrics  & D   & S  & M  & D  & S  & M   & D  & S  & M & D  & S  & M\\ \hline\hline
&\multicolumn{12}{c}{on \textit{Laval Indoor HDR Dataset}}\\\hline
RMSE$\downarrow$&0.181&0.218&0.207&0.202&0.274&0.275&0.170&0.213&0.207&\textbf{0.095}&\textbf{0.185}&\textbf{0.199}\\ 
si-RMSE$\downarrow$&0.055&0.147&0.155&0.065&0.147&0.160&0.048&0.143&0.151&\textbf{0.039} &\textbf{0.133} &\textbf{0.144}\\ \hline
&\multicolumn{12}{c}{on \textit{Web Dataset}}\\\hline
RMSE$\downarrow$& 0.758&0.682&0.609&0.511&0.443&0.435&0.522&0.492&0.455&\textbf{0.505}&\textbf{0.411}&\textbf{0.375}\\
si-RMSE$\downarrow$& 0.155&0.326&0.354&0.184&0.353&0.397&0.168&0.324&0.331&\textbf{0.134}&\textbf{0.285}&\textbf{0.301}\\
\hline
\end{tabular}

\end{table*}


\begin{figure*}[t]
    \centering
    \includegraphics[width=1.0\linewidth]{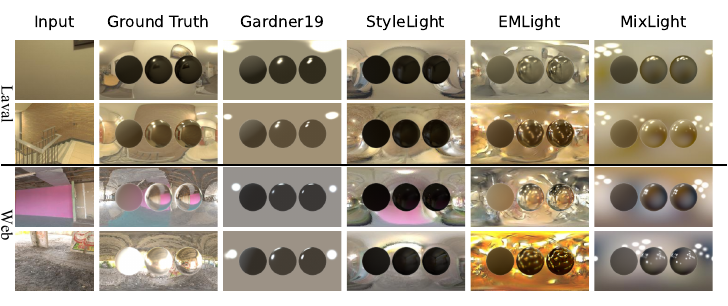}
    \caption{Visual comparison of the predicted results. In the first column are limited FOV images that serve as the input for all illumination estimation methods, followed by the predicted results visualized as illumination maps. Note that the first two rows are test samples from the Laval Indoor HDR Dataset, while the last two rows from the Web Dataset.}
    \label{fig:QualitativeLightmap}
\end{figure*}


\subsection{Quantitative Evaluation}

\noindent
In this section, MixLight is compared with three SOTA methods: Gardner19~\cite{gardner2019deep}, which solely relies on SG, StyleLight~\cite{wang2022stylelight}, known for direct illumination map generation, and EMLight~\cite{zhan2021emlight}, which utilizes SG to guide illumination map generation. The quantitative evaluation results on both the Laval Indoor HDR Dataset and the Web Dataset are presented in \Cref{tab:Quantitative}.

MixLight demonstrates superior performance compared to all SOTA methods across all evaluation metrics and ball materials. This is primarily attributed to the complementary advantages of SH and SG in the frequency domain, allowing for more accurate predictions.
Gardner19 simplifies the modeling of ambient light, leading to inaccuracies in predicting low-frequency illumination and subsequently affecting the RMSE metric. Additionally, the fixed number of SG light sources to 3 in Gardner19 oversimplifies the representation of high-frequency light sources, resulting in errors in the si-RMSE metric.
StyleLight exhibits notable degradation in the RMSE metric when transitioning from the Laval Indoor HDR Dataset to the Web Dataset, suggesting poor generalization performance due to overfitting to the limited-scale training set. Similarly, EMLight also suffers from overfitting errors, but to a lesser extent, attributed to the guidance of the low-dimensional SG model. However, EMLight does not consider the sparsity of the light source, leading to inaccurate predictions of high-frequency illumination.

Unlike the SG component in Gardner19 and EMLight, the sparse SG in MixLight effectively models the sparse real-life light sources due to the utilization of SLSparsemax, resulting in more precise high-frequency illumination predictions. In contrast to pure generation methods such as StyleLight, MixLight, as a low-dimensional parameterized model, exhibits less performance degradation from the Laval Indoor Dataset to the Web Dataset, highlighting its superior generalization capabilities.

\begin{figure*}[!th]
    \centering
    \includegraphics[width=1.0\linewidth]{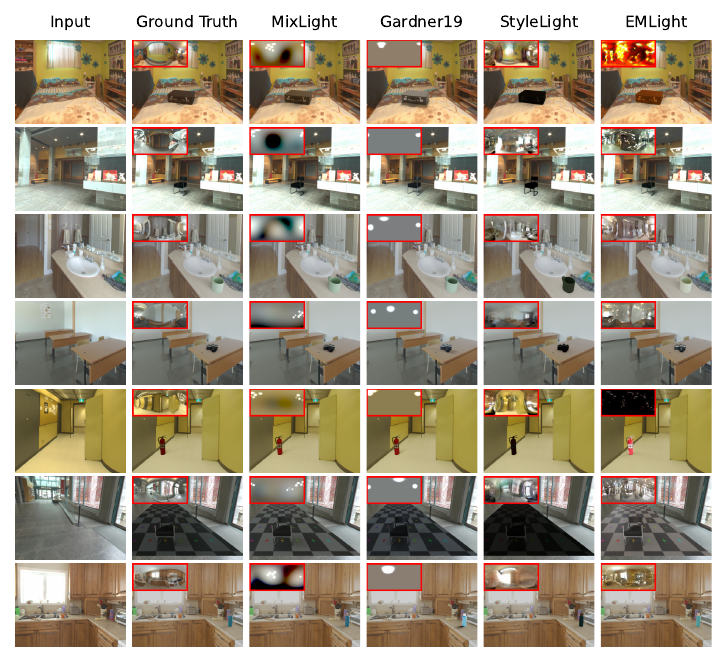}
    \caption{
    Visual comparisons of rendered results. In the first column are background images of 3D scenes, which act as the input for all illumination estimation methods. The predicted illumination maps (in the top-left red box of each rendered image) are then employed to render virtual objects.}
    \label{fig:QualitativeRendered}
\end{figure*}

\begin{figure}[!th]
    \centering
    \includegraphics[width=1.0\linewidth]{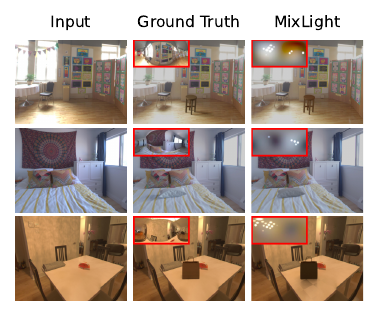}
    \caption{
    Failure cases in rendering results.}
    \label{fig:QualitativeRenderedFailure}
\end{figure}

\subsection{Qualitative Evaluation}

\noindent
This section visualizes the predicted results and the rendering results. 
Visualizing the predicted results allows direct observation of the light source's position and the ambient light, while visualized rendering results depict details (\eg, hard shadows) that are challenging to capture through quantitative evaluation but significantly impact realism. 
The SOTA methods compared are consistent with those used in the quantitative evaluation. 
Finally, some failure cases in rendering results will be shown and analyzed.

\textbf{Comparison of predicted results.} 
\Cref{fig:QualitativeLightmap} shows the predicted results of various methods, along with the GT illumination maps. The first two rows are test samples from \textit{Laval Indoor HDR Dataset}, with the last two rows from the \textit{Web Dataset}. 
In all four samples, the light source positions and ambient color variance predicted by MixLight are all close to the GT. In contrast, Gardner19 oversimplifies light sources and ambient light representation by fixing 3 light sources and one ambient light color, leading to a poor fit for the strip lights in the first sample and an inability to represent the free-varying ambient light as demonstrated by MixLight. StyleLight and EMLight tend to overfit the indoor style of the Laval Indoor HDR Dataset, performing well on the first two samples but predict inaccurately when the samples are collected from the internet with a significantly different style, as observed in the third and fourth rows. For instance, in the third sample, the pink walls, and in the fourth sample, the abandoned house are not present in the Laval Indoor HDR Dataset. In these two cases, StyleLight and EMLight still generate ceiling, yellow walls, and exquisite doors and windows consistent with the Laval Indoor HDR Dataset style, which is notably different from the GT. Additionally, EMLight does not constrain the sparsity of SG, resulting in the presence of numerous unnecessary light sources in the generated illumination maps.

It is important to note that the comparison of visualized illumination maps serves as a supplementary tool for understanding the limitations of certain methods as identified in the quantitative evaluation section (e.g., Gardner19 oversimplifies the light source representation).
The predicted results of low-dimensional parametric methods seem to be smoother.
However, the perceived smoothness or sharpness of the LDR illumination map does not inherently reflect the accuracy of the prediction. This is due to two primary factors:
(1) The angular information of light is convolved and highly blurred during rendering. Thus the perceived smoothness or sharpness might not be important because it will inevitably be blurred after rendering.
(2) Typically, the subjective quality in LDR space does not necessarily align with the results in HDR. In the HDR space, the ambient light predicted by all methods appears smooth (compared with the light source).


\textbf{Comparison of rendered results.} 
\Cref{fig:QualitativeRendered} shows the rendered images alongside the corresponding predicted results used for rendering. It can be observed that MixLight predicts accurately in terms of light source position and variation of ambient light, thus producing realistic rendering results where the color, brightness, and shadows of the virtual objects are close to the GT. 
Conversely, Gardner19 oversimplifies the modeling of ambient light and light sources. The use of a single color fails to accurately represent the ambient light, resulting in inaccurate color and brightness of virtual objects after rendering, especially in the first sample. Additionally, the limited number of three light sources struggles to represent the variable light sources in the real world, leading to overly harsh shadows in the rendering results, particularly noticeable in the last sample.
StyleLight tends to overfit the Laval Indoor HDR Dataset, and its brightness predictions are generally lower than the ground truth, resulting in excessively low brightness in the rendering results.
In contrast, EMLight employs the predicted parameters to guide the generation of illumination maps, thereby mitigating the overfitting issue to some extent. This approach demonstrates improved performance in the second, third, and fourth samples. However, the generative model trained on a small-scale dataset still exhibits instability, leading to significant distortion of the illumination maps in the first sample and resulting in orange-colored objects in the rendering.

\textbf{Failure cases in rendering results.} Illumination estimation is still a difficult problem due to its \textbf{strong ill-posed nature}. 
So even in other papers, it is normal for the proposed method to have limited effectiveness on some samples. The proposed MixLight exceeds all SOTAs in its entirety but also fails on individual samples, as illustrated in \Cref{fig:QualitativeRenderedFailure}.
There are two main types of failure: inaccurate brightness or location prediction of light sources.
For one thing, the brightness of the light source has greater uncertainty than other parameters, whose loss is more difficult to converge during training. 
For the chair and pillow in the first two examples in \Cref{fig:QualitativeRenderedFailure}, the brightness is predicted too weak to render clear shadows even if the light source position is very accurate. For the bag in the third example, the brightness is predicted too high, leading to harder Shadows.
For another, the position of light Sources could be predicted better if more clues (\eg, shading and shadows) exist in the input. Without clues, even humans can not guess accurate positions. 
The third example has a huge desktop with neither shading nor clear shadows, where the shadow indicates light from the top but is insufficient to help further locate its azimuth.


\begin{table*}[t]
  \centering
    \caption{Ablation study of choosing different representation models for ambient light and light sources. There are four combinations in total, namely using SH and SG to represent ambient light and light sources (MixLight), using SG for both ambient light and light sources (SGG), using SH for both ambient light and light sources (SHH), and using SG for ambient light and SH for light sources (SGH).}
  \label{tab:ablationRepresentationModels}
\begin{tabular}{c|ccc|ccc|ccc|ccc}
\hline
\multirow{2}{*}{Metrics} & \multicolumn{3}{c|}{MixLight} & \multicolumn{3}{c|}{SGG} & \multicolumn{3}{c|}{SHH}& \multicolumn{3}{c}{SGH}\\ \cline{2-13} 
 & D   & S  & M  & D  & S  & M   & D  & S  & M & D  & S  & M\\ 
 \hline \hline 
&\multicolumn{12}{c}{on \textit{Laval Indoor HDR Dataset}}\\
 \hline 
RMSE$\downarrow$&\textbf{ 0.095 }&\textbf{ 0.185 }&\textbf{ 0.199 }&0.121&0.223&0.230&0.097&0.210&0.232&0.105&0.212&0.233 \\
si-RMSE$\downarrow$ &\textbf{ 0.039 }&\textbf{ 0.133 }&\textbf{ 0.144 }&0.067&0.179&0.185&0.053&0.158&0.165&0.057&0.159&0.165 \\
\hline
&\multicolumn{12}{c}{on \textit{Web Dataset}}\\
 \hline 
RMSE$\downarrow$&\textbf{ 0.505 }&\textbf{ 0.411 }&\textbf{ 0.375 }&0.585&0.539&0.507&0.525&0.463&0.425&0.506&0.448&0.415 \\
si-RMSE$\downarrow$&\textbf{ 0.133 }&\textbf{ 0.285 }&\textbf{ 0.301 }&0.226&0.445&0.448&0.213&0.403&0.396&0.200&0.391&0.387 \\
\hline
\end{tabular}
\end{table*}


\begin{table*}[!th]
  \centering
    \caption{Ablation study of the proposed SLSparsemax used in MixLight.
  The MixLight variants with SLSparsemax (SLS), original sparsemax (S), and No Sparsemax (NS) undergo quantitative testing on two datasets. \textbf{Best} scores are presented in bold, while \textit{second-best} scores are displayed in italics.}
  \label{tab:ablationSLSparsemax}
\begin{tabular}{c|ccc|ccc|ccc}
\hline
\multirow{2}{*}{Metrics} & \multicolumn{3}{c|}{No Sparsemax (NS)}& \multicolumn{3}{c|}{Sparsemax (S)} & \multicolumn{3}{c}{SLSparsemax (SLS)} \\ 
\cline{2-10} 
 & D   & S  & M & D & S  & M  & D  & S  & M  \\ 
 \hline \hline 
&\multicolumn{9}{c}{on \textit{Laval Indoor HDR Dataset}}\\
 \hline 
RMSE$\downarrow$&0.111&\textit{0.198}&0.218& \textit{0.102}&0.199&\textit{0.209}&\textbf{0.095}&\textbf{0.185}&\textbf{0.199}\\ 
si-RMSE$\downarrow$&0.052& \textit{0.141}&0.151&\textit{0.043}&\textit{0.141}&\textit{0.150}&\textbf{0.039}&\textbf{0.133}&\textbf{0.144}\\
\hline
&\multicolumn{9}{c}{on \textit{Web Dataset}}\\
 \hline 
RMSE$\downarrow$&0.527&0.436&0.396&\textit{0.510}&\textit{0.432}&\textit{0.384}&\textbf{0.505}&\textbf{0.411}&\textbf{0.375}\\ 
si-RMSE$\downarrow$&0.159&0.307&0.326&\textit{0.144}&\textit{0.305}&\textit{0.314}& \textbf{0.134} &\textbf{0.285} &\textbf{0.301}\\
\hline
\end{tabular}
\end{table*}

\begin{figure*}[!ht]
    \centering
    \subfloat[GT]{\includegraphics[width=0.24\linewidth]{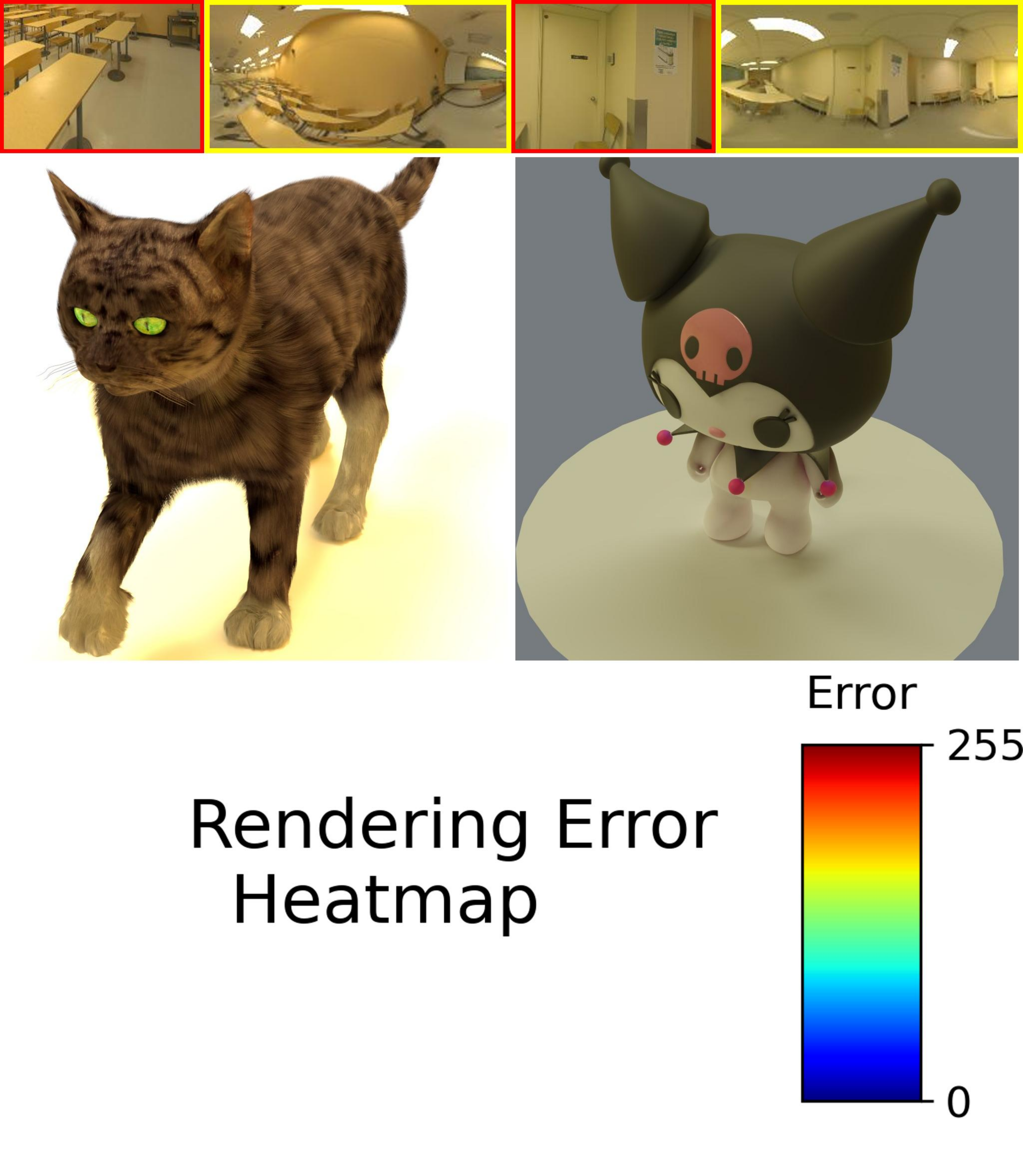}}
    \hfill
    \subfloat[NS]{\includegraphics[width=0.24\linewidth]{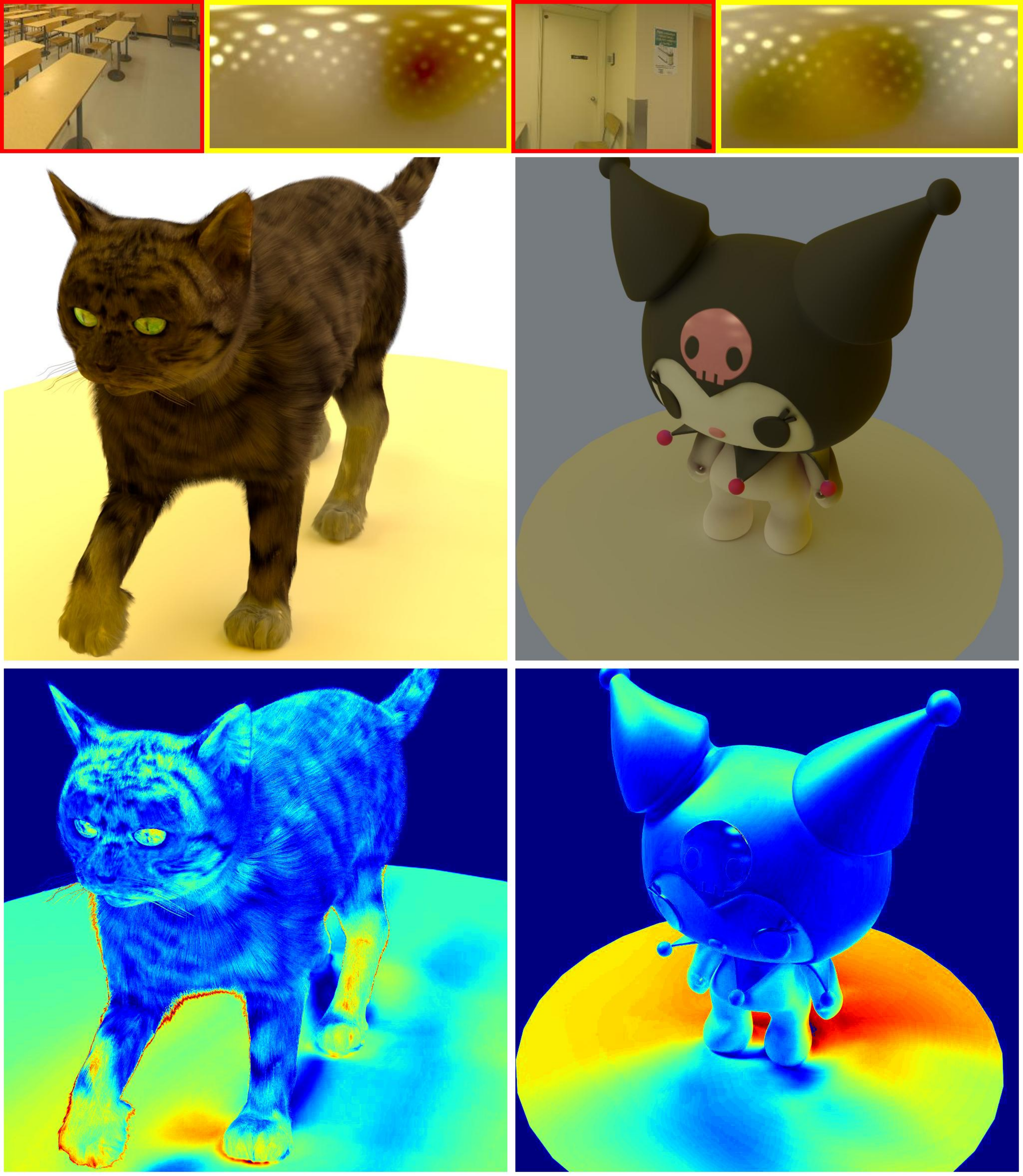}}
    \hfill
    \subfloat[S]{\includegraphics[width=0.24\linewidth]{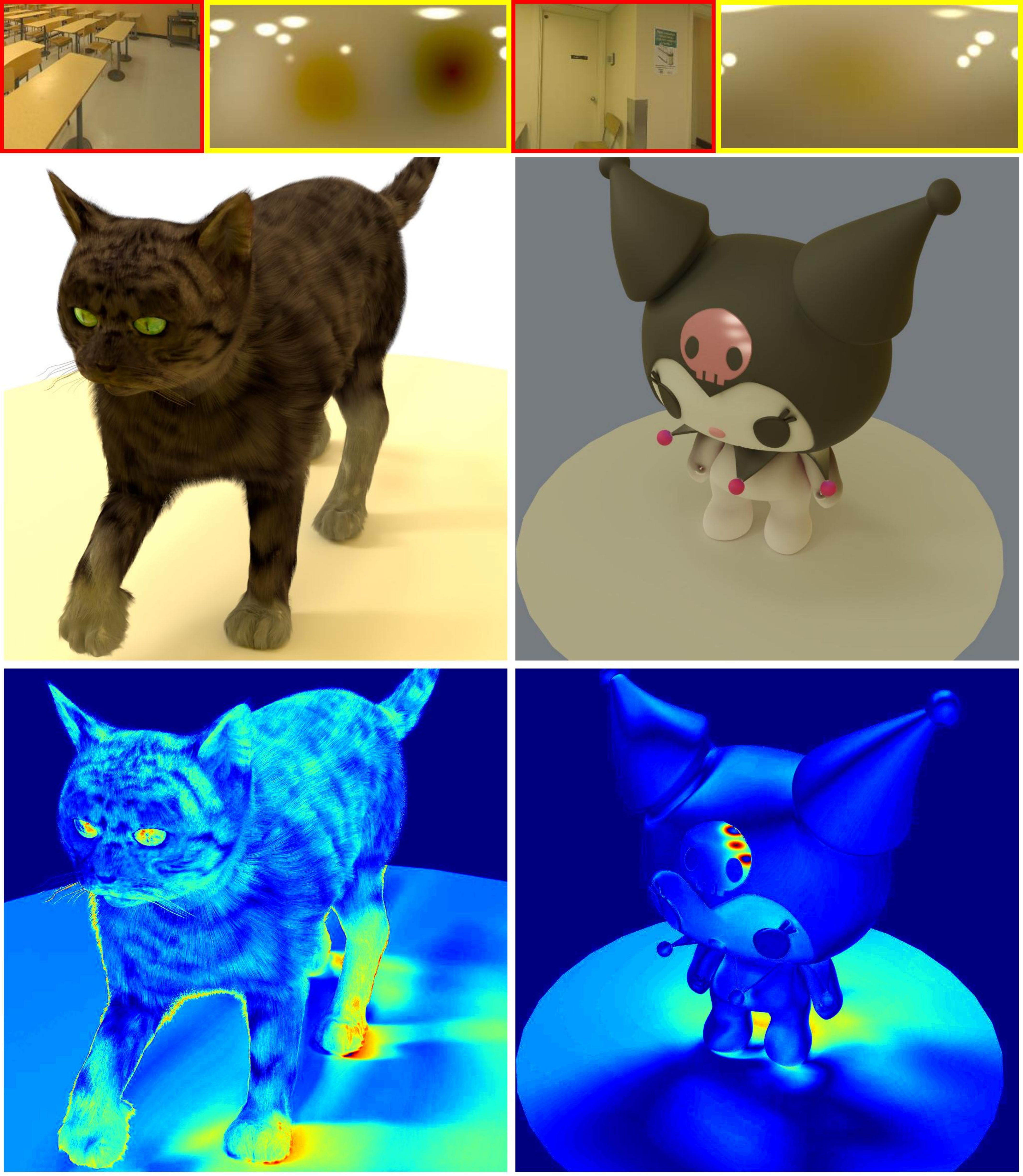}}
    \hfill
    \subfloat[SLS]{\includegraphics[width=0.24\linewidth]{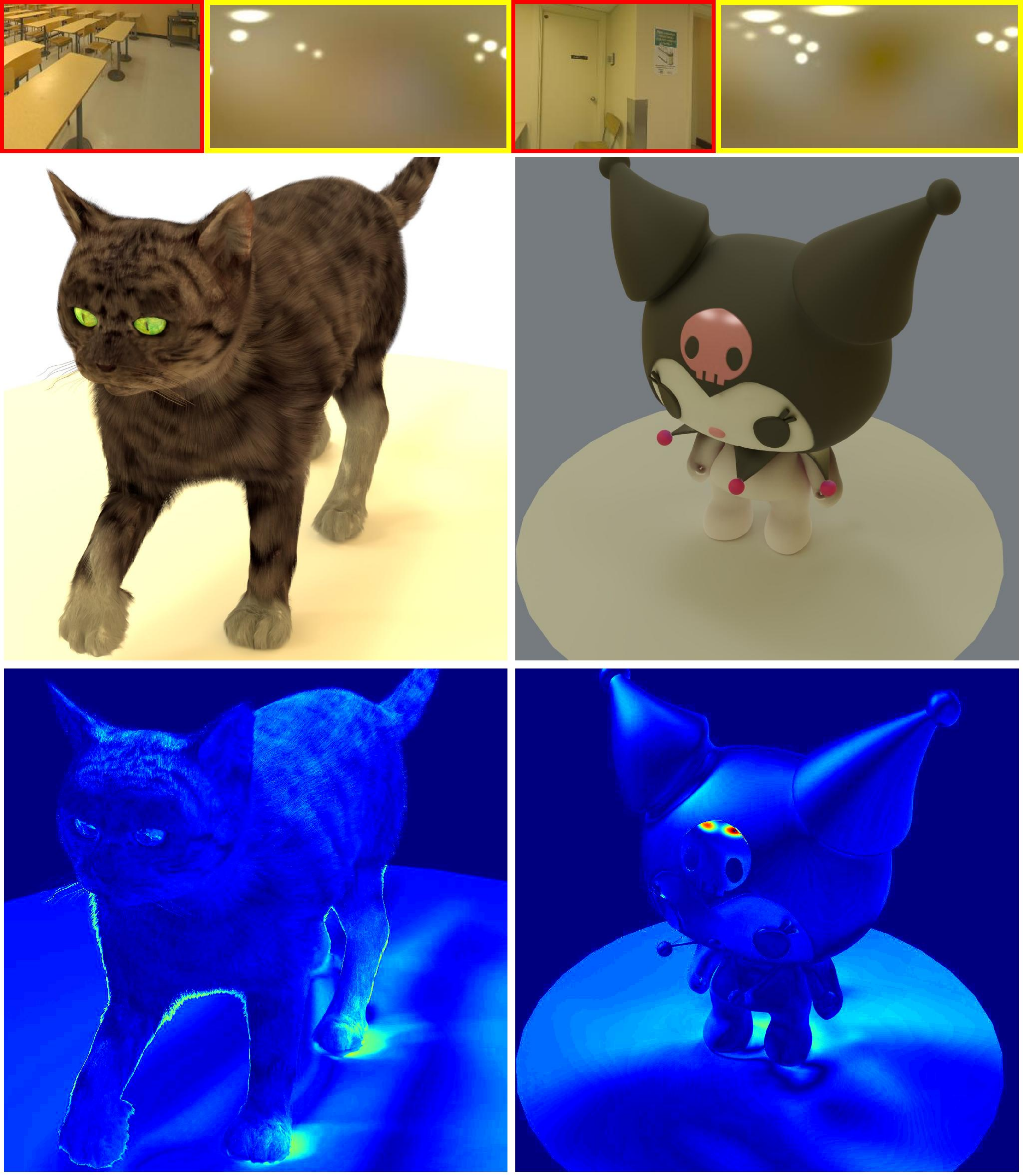}}
    \caption{Visual comparison of the rendered results from MixLight with No Sparsemax (NS), original Sparsemax method (S), or SLSparsemax (SLS). In the top-left red box of each example is the input limited FOV image, while in the top-right yellow box is the predicted illumination map.}
    \label{fig:ablationSLSparsemax}
\end{figure*}

\subsection{Ablation Study}

\noindent
MixLight uses SH to represent ambient light and SG to represent light sources.
\Cref{tab:ablationRepresentationModels} shows the ablation study of choosing different representation models for ambient light and light sources, using the same datasets, metrics, and comparison methods as quantitative evaluation. 
The other three (model selection) combinations are SG for both ambient light and light sources (SGG), SH for both ambient light and light sources (SHH), and SG for ambient light but SH for light sources (SGH).
Note that these three combinations have the same or similar parameter scales with MixLight, and more details about the parameter settings are in the supplementary file.
As shown in \Cref{tab:ablationRepresentationModels}, MixLight surpasses the other three combinations in all test sets, all materials of the spheres, and all metrics, mainly benefiting from the accurate prediction ability of SH and SG to low-frequency and high-frequency information respectively.

Another experiment is conducted to quantitatively compare the MixLight variants with SLSparsemax (SLS), original sparsemax (S), and No Sparsemax (NS), as depicted in \Cref{tab:ablationSLSparsemax}. 
\textbf{Best} scores are presented in bold, while \textit{second-best} scores are displayed in italics. 
Among the 12 metric results in the two test sets, SLS outperforms the original S with most best scores while S gets more second-best scores than NS.
This decreasing error from NS, S to SLS can be observed more intuitively in visualized rendering results and respective error maps in \Cref{fig:ablationSLSparsemax}, where S and SLS achieve a much sparser light source prediction than NS, better at mimicking the sparse light sources in GT and consequently producing pronounced shadows. 
SLSparsemax further generates locally clustered yet globally sparse light sources that closely resemble reality, thus rendering more realistic shadows and highlights and getting better scores in \Cref{tab:ablationSLSparsemax}.

\section{Limitations and Future Works}
\noindent
This paper explores the feasibility of achieving a more accurate illumination representation model by combining inherent representation models. By observing the law of light sources in indoor scene datasets, we propose the sparsity assumption of light sources, and further design SLSparsemax modules based on this assumption. However, it should be noted that there are also times when the sparsity assumption does not apply. Illumination estimation for outdoor scenes is also not the focus of this paper for the time being. In addition, predicting spatially-varying illumination and predicting illumination with more details are all directions that can be further explored in the future.

\textbf{Assumption of sparsity.} In this paper, an assumption of sparsity is made: in a scene, there are usually only a few light sources, and the number of light sources is variable. This assumption is made as the target dataset (Laval Indoor) presents such sparsity characteristics. It should be noted that sometimes this assumption does not hold (\eg, multiple small light sources in the laboratory). Despite this limitation, making this assumption is a step forward from previous assumptions explicitly or implicitly used by previous SG methods. For example, Gardner19 \cite{gardner2019deep} uses only three SG kernels, and \cite{weber2022editable} uses only one SG kernel to represent light sources. In the future, SLSparsemax could be introduced to more SG works to model sparse and variable numbers of SG light sources. A more general light source assumption can also be proposed based on the current one.

\textbf{Outdoor scenes.} Outdoor scenes have not been considered in this paper. However, both SH and SG can be promoted to outdoor scenes like~\cite{cheng2018learning,dastjerdi2023everlight}. Therefore, as a SH, SG combined method, MixLight can also be promoted to predict outdoor illumination. Additionally, outdoor light sources (typically the sun) are fewer in number but have higher energy, different from indoor light sources. This difference previously blocks the formation of a unified indoor-outdoor light source model. However, this bottleneck is naturally suitable for the proposed SLSparsemax to tackle, which produces number-variant light sources adaptively. It can be further investigated in future work.

\textbf{Spatially-varying illumination.} MixLight is currently not a spatially-varying illumination representation. Considering that the depth information of scenes can be obtained from the Laval indoor HDR Dataset, this problem could be addressed by generating spatially-varying training pairs when pre-processing data, similar with~\cite{garon2019fast}, or by regressing the depth of light sources and promoting them from 2D sphere surface to 3D like~\cite{gardner2019deep}. 

\textbf{Generalization performance.} Recovering photo-realistic scene details from a single limited FOV image with a lack of mirror objects remains a challenge, as previous GAN-based approaches have shown poor generalization performance. Inputting multiple images can provide more scene details, which may be a possible solution to address this challenge. Building a large-scale real HDR dataset also helps improve the generalization performance.

\section{Conclusions}
\noindent
This paper introduces MixLight, a joint illumination representation model that leverages the complementary advantages of SH and SG in the frequency domain. 
SLSparsemax is designed to help MixLight achieve as sparse light source predictions as that in real life. 
The results of quantitative and qualitative evaluations demonstrate that MixLight accurately predicts illumination from single indoor scene limited FOV images. 
Further experiments on a Web Dataset show that parametric methods have better generalization performance compared to generation methods.

\clearpage
\setcounter{page}{1}

\appendices

\section{SH-SG comparison experiment}\label{DetailsOf}
\noindent
In the \textit{Introduction} section of the main manuscript, an experiment (Fig. 1) is used to illustrate the different advantages of SH and SG in representing ambient light and light sources. The training and testing sets used are both from the Laval Indoor HDR Dataset~\cite{gardner2017learning}. The partition method of the dataset is consistent with the experimental settings in the main manuscript. The test set consisted of 200 randomly selected samples. For each sample, ambient light (or light sources) is predicted and used to render three balls. 200 samples render 600 results. For each rendering result, the error is calculated by the formula $\sqrt{\alpha\cdot\beta}$, where $\alpha$ represents the Root Mean Square Error (RMSE) and $\beta$ represents si-RMSE~\cite{grosse2009ground}. Finally, 600 error values are presented in the form of a violin chart (\eg, Fig. 1a shown in the main manuscript), which shows the distribution of the error values on the test set, when using one of the SH and SG models to predict ambient light or light source.

To fairly compare the performance of SH or SG on the ambient light prediction task or the light source prediction task, different versions of SG are designed to predict light sources and ambient light.

In the \textit{ambient light estimation task}, a 9-kernel \textbf{smooth} SG is designed, with its angular size $s$ adjusted to 0.2423, which ensures that the 9 SG functions cover the sphere as comprehensively as possible while maintaining a decay of roughly 0.5 on the boundaries. This smooth version of SG, along with its $s$ adjusting method, mirrored the approach in \cite{li2019deep}. The only difference is that the number of SG kernels is reduced to 9 in this experiment to keep it focused on capturing low-frequency ambient light. This also helps to maintain SG's parameter number at 27 (in RGB three channels), consistent with a 2nd order SH served as a competitor in ambient light prediction task.

In the \textit{light source estimation task}, a \textbf{sharp} version of SG is utilized, which is essentially the spherical Gaussian distribution proposed in EMLight. 
The angular size $s$ is set to 0.0025 to keep the SG kernel function as sharp as the light sources.
Then compare the sharp 128-kernel SG with a 6th-order SH, which comprised 132 and 147 parameters, respectively. While the parameters of SG and SH for the light sources prediction task were not entirely consistent, the 6th-order SH was the closest in terms of parameter scale with 128-kernel sharp SG.

\section{Original Sparsemax}
\noindent
Sparsemax~\cite{martins2016softmax} operates under the assumption that although the network's output distribution values are fundamentally reliable after numerous training iterations, they are still not sparse enough. To address this, it sorts all elements within the input distribution value in descending order and retains only the leading large values, effectively filtering out small values that are close to zero. Additionally, to minimize alterations to the somewhat trusted input, it only applies an overall offset to the distribution value to preserve the distinction between elements, while striving to retain as many elements as possible to minimize changes in quantity.




\section{Ablation Study details}
\noindent
Comparison fairness is ensured when comparing MixLight with 3 other combinations in the ablation study. 

In the first experiment of the ablation study, we replace the 2nd order SH of MixLight with 9-kernel SG (smooth version SG) to construct SGG. We replace the 128-kernel SG of MixLight with 6th order SH to construct SHH. We replace both above to construct SGH. So MixLight, SGG, SHH, and SGH have the same or close (27+132), (27+132), (27+147), and (27+147) parameters respectively.

\section{More results}

\begin{figure*}[!t]
    \centering
    \includegraphics[width=1.0\linewidth]{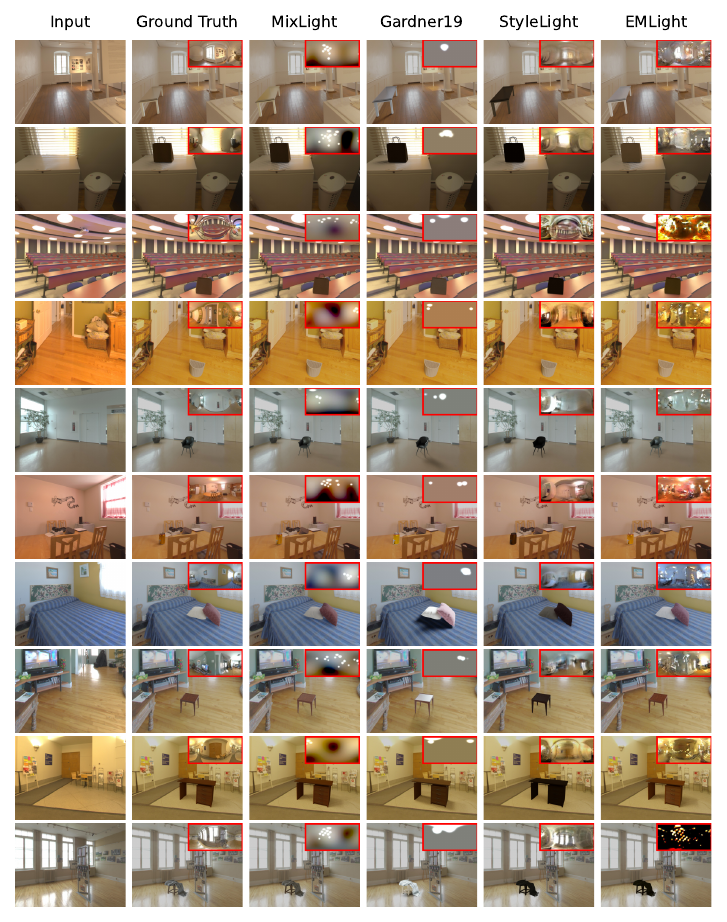}
    \caption{More Visual comparisons of rendered results. In the first column are background images of 3D scenes, which act as the input for all illumination estimation methods. The predicted illumination maps (in the top-right red box of each rendered image) are then employed to render virtual objects.}
    \label{fig:moreResultsRendered}
\end{figure*}

More rendered results are given in \Cref{fig:moreResultsRendered} to supplement Fig. 6 of the main manuscript.

\begin{figure*}[!t]
    \centering
    \includegraphics[width=1.0\linewidth]{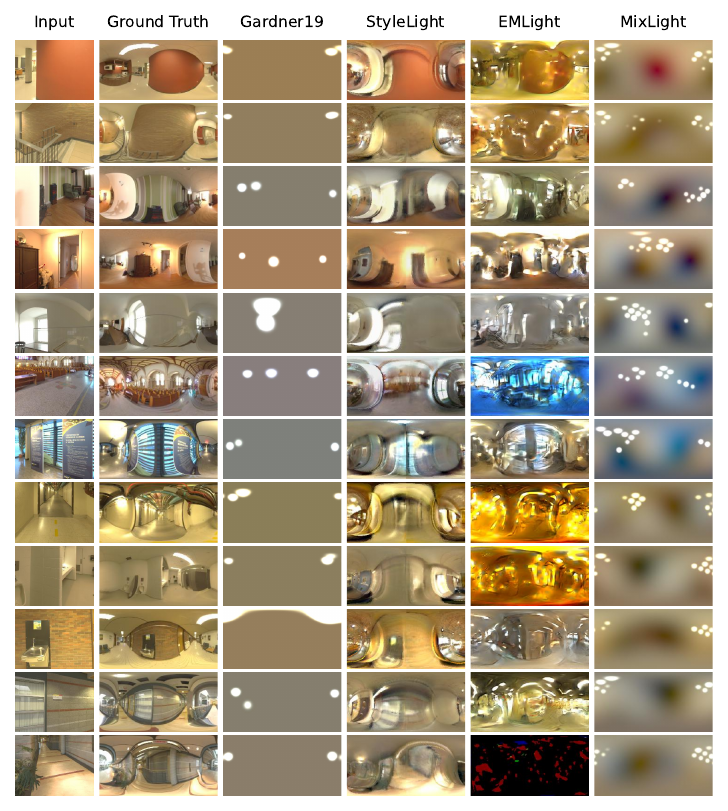}
    \caption{More visual comparison of the predicted results on Laval Indoor HDR Dataset~\cite{gardner2017learning}. These samples are all randomly sampled from the test set of the Laval Indoor HDR Dataset, which is the most frequently used dataset of the illumination estimation academy. The MixLight proposed in this paper and three other compared SOTAs are all trained based on the training set from Laval Indoor HDR Dataset. Therefore, all the above illumination prediction neural networks are familiar with the style of this dataset and could get better scores when tested on the test set of the Laval Indoor HDR Dataset (see Table 1 in the main manuscript).}
    \label{fig:moreResultsOnLaval}
\end{figure*}

\begin{figure*}[!t]
    \centering
    \includegraphics[width=1.0\linewidth]{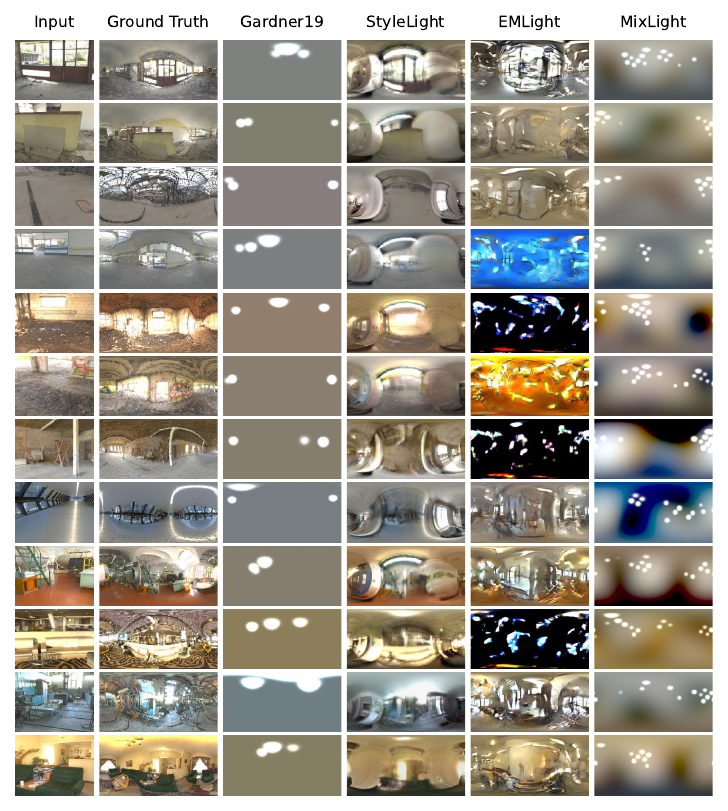}
    \caption{More visual comparison of the predicted results on Web Dataset. These samples are all randomly sampled from the test set of the Web Dataset. The Web Dataset is collected from the internet, has no intersection with the Laval Indoor HDR Dataset, and thus has diverse styles that have never been seen by all compared methods. When tested on this unfamiliar ``in the wild'' test set, the performance of all methods degenerates to varying degrees, as can be seen in Table 1 of the main manuscript. The degradation is most severe in the generative approach, as they always reproduce the style of the old Laval Indoor HDR Dataset on this new test set. Specifically, more failed predictions have emerged for EMLight, such as the fifth and seventh samples. }
    \label{fig:moreResultsOnWeb}
\end{figure*}

In Fig. 5 of the main manuscript, the predicted results are visualized as illumination maps to help analyze the pros and cons of various methods given in quantitative evaluation, where rendering results is the final goal that needs to be evaluated quantitatively. To aid the assessment, here are more direct predictions visualized in the form of light maps. Again, the results are split into two different test sets to compare how different methods behave on test sets from familiar (\Cref{fig:moreResultsOnLaval}) to unfamiliar styles (\Cref{fig:moreResultsOnWeb}).

\ifCLASSOPTIONcaptionsoff
  \newpage
\fi

\bibliographystyle{IEEEtran}
\bibliography{main}

\end{document}